\documentclass[10pt,twocolumn,letterpaper]{article}

\usepackage[pagenumbers]{cvpr}      %
\usepackage[dvipsnames]{xcolor}
\usepackage{float}

\usepackage{graphicx}
\usepackage{amsmath}
\usepackage{amssymb}
\usepackage{booktabs}
\usepackage[symbol]{footmisc}

\usepackage{cancel}

\usepackage{svg}

\usepackage{tikz}
\usetikzlibrary{positioning}
\usetikzlibrary{shapes.geometric}

\usepackage{bm}

\usepackage{breqn}
\usepackage{nicematrix}
\usepackage{booktabs}
\usepackage{makecell}
\usepackage{multirow}
\usepackage{colortbl}

\usepackage[accsupp]{axessibility}  %

\usepackage[noend,ruled,linesnumbered]{algorithm2e}
\SetKwInput{KwInput}{Input}                %
\SetKwInput{KwOutput}{Output}              %
\SetKwInput{KwInitialize}{Initialize}      %

\usepackage{amsthm}

\usepackage{array}
\newcolumntype{L}[1]{>{\raggedright\let\newline\\\arraybackslash\hspace{0pt}}m{#1}}
\newcolumntype{C}[1]{>{\centering\let\newline\\\arraybackslash\hspace{0pt}}m{#1}}
\newcolumntype{R}[1]{>{\raggedleft\let\newline\\\arraybackslash\hspace{0pt}}m{#1}}

\usepackage[pagebackref,breaklinks,colorlinks]{hyperref}
\hypersetup{
    colorlinks=true,
    allcolors= Blue, %
    pdftitle={},
    pdfpagemode=FullScreen,
    }

\SetKwComment{Comment}{\color{RoyalBlue}$\triangleright$\ }{}

\usepackage[capitalize,nameinlink]{cleveref}
\crefdefaultlabelformat{#2\textbf{#1}#3}
\crefname{equation}{}{}
\creflabelformat{equation}{#2\textup{\bf #1}#3}
\crefname{equation}{Eq.}{Eqs.}
\Crefname{equation}{Equation}{Equations}
\crefname{figure}{Fig.}{Figs.}
\Crefname{figure}{Figure}{Figures}
\crefname{table}{Tab.}{Tabs.}
\Crefname{table}{Table}{Tables}
\crefname{section}{Sec.}{Secs.}
\Crefname{section}{Section}{Sections}
\crefname{problem}{Problem}{Problems}
\Crefname{problem}{Problem}{Problems}
\crefname{definition}{Definition}{Definitions}
\Crefname{definition}{Definition}{Definitions}
\crefname{lemma}{Lemma}{Lemmas}
\Crefname{lemma}{Lemma}{Lemmas}
\crefname{theorem}{Thm.}{Thms.}
\Crefname{theorem}{Theorem}{Theorems}
\crefname{remark}{Rmk.}{Rmks.}
\Crefname{remark}{Remark}{Remarks}
\crefname{enumi}{item}{items}
\Crefname{enumi}{Item}{Items}
\crefname{algocf}{Alg.}{Algs.}
\Crefname{algocf}{Algorithm}{Algorithms}
\crefname{assumption}{Asm.}{Asms.}
\Crefname{assumption}{Assumption}{Assumptions}
\crefname{ALC@unique}{line bla}{lines}
\Crefname{ALC@unique}{Line bla}{Lines}

\def\cvprPaperID{10928} %
\def\confName{ICCV}
\def\confYear{2023}

\usepackage{soul}
\usepackage{enumitem}
\usepackage{esvect}
\usepackage{array}
\usepackage{makecell}
\newcolumntype{L}{>{\centering\arraybackslash}m{2cm}}

\MakeRobust{\vv}

\usepackage[colorinlistoftodos]{todonotes}

\newcommand{\paragraphCVPR}[1]{\vspace{6pt}\noindent{\bf #1:}}

\usepackage{layout}

\begin{document}

\title{%
Robust Frame-to-Frame Camera Rotation Estimation in Crowded Scenes
} %

\author{
Fabien Delattre $^1$\thanks{work done while intern at MERL} \and
David Dirnfeld $^1$ \and
Phat Nguyen $^1$ \and
Stephen Scarano $^1$ \and
Michael J. Jones $^2$ \and
Pedro Miraldo $^2$ \and
Erik Learned-Miller $^1$  \and \\
$^1$ University of Massachusetts Amherst \ \ \
$^2$ Mitsubishi Electric Research Laboratories (MERL)
\ \ \
}

\maketitle

\begin{abstract}
   We present an approach to estimating camera rotation in crowded, real-world scenes from handheld monocular video. While camera rotation estimation is a well-studied problem, no previous methods exhibit both high accuracy and acceptable speed in this setting.  Because the setting is not addressed well by other datasets, we provide a new dataset and benchmark, with high-accuracy, rigorously verified ground truth, on 17 video sequences. Methods developed for wide baseline stereo (e.g., 5-point methods) perform poorly on monocular video. On the other hand, methods used in autonomous driving (e.g., SLAM) leverage specific sensor setups, specific motion models, or local optimization strategies (lagging batch processing) and do not generalize well to handheld video. Finally, for dynamic scenes, commonly used robustification techniques like RANSAC require large numbers of iterations, and become prohibitively slow. We introduce a novel generalization of the Hough transform on $SO(3)$ to efficiently and robustly find the camera rotation most compatible with optical flow. Among comparably fast methods, ours reduces error by almost 50\% over the next best, and is more accurate than any method, irrespective of speed. This represents a strong new performance point for crowded scenes, an important setting for computer vision. The code and the dataset are available at \href{https://fabiendelattre.com/robust-rotation-estimation}{https://fabiendelattre.com/robust-rotation-estimation}. %
\end{abstract}

\section{Introduction}
\label{sec:intro}
The estimation of camera motion through a scene is a fundamental problem in computer vision that is highly related to a number of vision tasks such as motion segmentation~\cite{Bideau2016ItsMA}, video stabilization~\cite{Liu2013VideoStabilization}, 3D reconstruction~\cite{Debevec3DReconstruction}, visual odometry~\cite{Nistr2004VO}, Simultaneous Localisation and Mapping (SLAM)~\cite{MurArtal2015ORBSLAMAV}, Structure-from-Motion (SfM)~\cite{schoenberger2016sfm}, human-computer interaction~\cite{Newcombe2011KinectFusion}, autonomous navigation~\cite{Annett2012RobotNavigation}, and many more. Hence, developing a method that can accurately predict the camera's movement through a scene is critical in solving these problems.

\begin{figure}[t]
  \centering
     \includegraphics[width=1\linewidth]{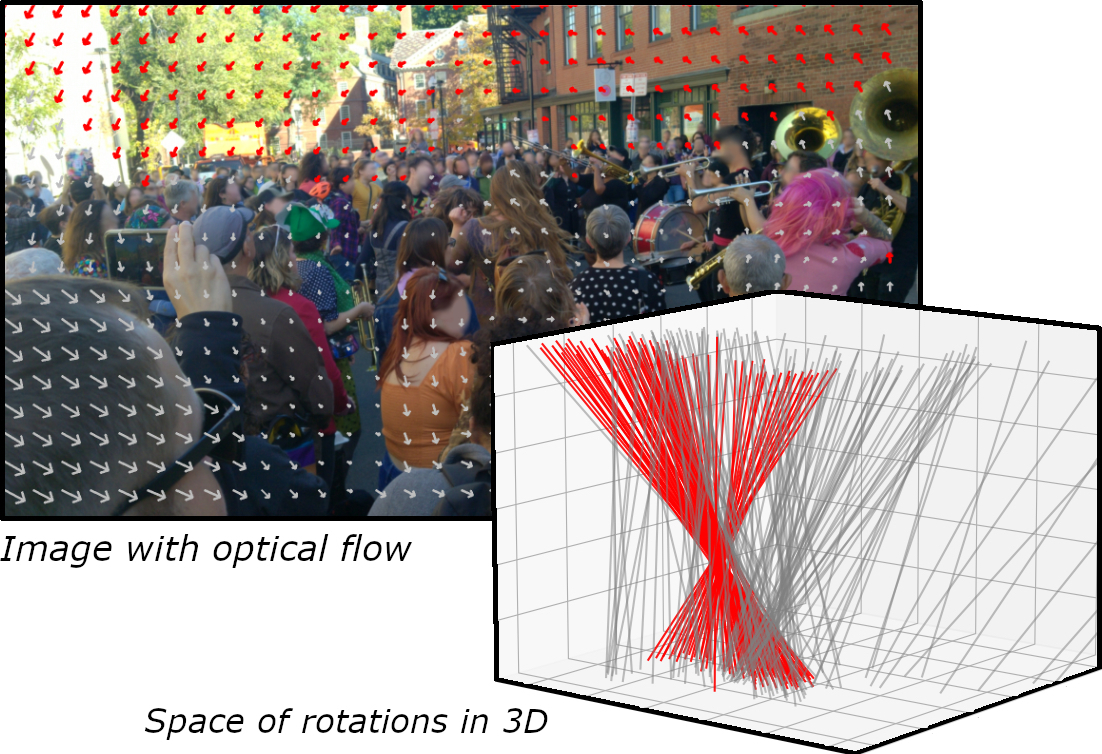}
    \caption{{\bf Left.} A frame from our BUSS dataset of crowded scenes. The red vectors show optical flows compatible with the winning rotation estimate $R^*$, indicating the rotation of the camera. Gray vectors show optical flows not explained purely by $R^*$. 
    {\bf Right.} The three axes show the space of rotations in 3D. Each line shows the one-dimensional set of rotations that are compatible with a single optical flow vector. The red lines (corresponding to the red flow vectors in the top figure) intersect in a single small bin, indicating that their optical flows are compatible with the same rotation. The gray lines, which are affected by other motion effects, are scattered in an unstructured manner, and correspond to the gray optical flows above. Our algorithm finds the set of lines with greatest coherence in $SO(3)$, revealing the rotation $R^*$ of the camera.}
   \label{fig:intro}
\end{figure}

As the camera moves through the scene, the motion field depends not only on the camera's motion but also on the scene's geometry and objects' motion in the environment. Given a sufficiently crowded location with many moving objects (e.g., pedestrians and vehicles), estimating the camera motion requires the difficult task of distinguishing between static and moving objects.
This paper proposes a novel, robust method of estimating camera rotation in crowded scenes such as the one shown in \cref{fig:intro}.

It is important to clarify the difference between frame-to-frame camera motion estimation and relative pose estimation. Specifically, camera motion estimation is a constrained version of relative pose estimation where only two views are used, constrained to be (a) spatially close, (b) temporally close, and (c) taken from the same camera, which matches the case of adjacent frames in a moving-camera video.

Nowadays, many authors focus on relative pose estimation using point correspondences. Most of these methods focus on estimating the essential matrix~\cite{LonguetHiggins1981ACA,Nistr2004VO},
which works best in the presence of large parallax~\cite[Remark 5.2]{ma2004invitation} (large baselines). Therefore, correspondence-based methods are primarily used for offline localization and mapping strategies such as SfM and 3D reconstruction, or online pipelines with local optimization like SLAM. In contrast, optical flow-based methods are better suited for small motions, which is the domain of interest in this paper.

As in state-of-the-art correspondence-based relative pose problems \cite{Kneip2013DirectOO}, the best optical flow-based methods for frame-to-frame camera motion estimation focus on decoupling the transformation into rotation- and translation-only estimation~\cite{Bruss1982PassiveN,Bideau2016ItsMA}. While there are fast and accurate solutions to motion estimation, they are highly sensitive to moving objects in the scene--they frequently break down with significant numbers of moving objects in the scene. 
Similarly to correspondence-based techniques, optical flow-based methods are often used within RANSAC~\cite{Fischler1981RANSAC} to handle locally wrong optical flow and moving objects, and thereby increase robustness. In this paper, we focus on rotation estimation since flow-based translation estimation given rotation estimates can be easily computed as shown in~\cite{Bruss1982PassiveN,Bideau2016ItsMA}.

We propose a new method to estimate the camera rotation based on optical flow. Our approach can be used for highly dynamic scenes, from the assumption that optical flow from faraway points is less sensitive to dynamic objects in the scene. The proposed technique uses a compatible rotation voting mechanism and does not require RANSAC (see Fig.~\ref{fig:intro}).
In addition, since public datasets only contain static scenes or have minor dynamic objects (a large portion of the frames contain static environments), we acquire a new and challenging dataset of 17 sequences in (anonymized) crowded environments. The dataset will be made available. To summarize, our contributions are as follows:
\begin{itemize}
  \item[--] A novel robust frame-to-frame camera rotation estimation algorithm based on optical flow that finds compatible rotations using a voting mechanism based on the Hough transform in the space of 3D rotations; 
  \item[--] We show that our algorithm significantly outperforms the discrete and continual baselines in highly dynamic
  scenes and performs comparably in static scenes; and
  
  \item[--] We provide a new dataset of highly dynamic scenes called BUsy Street Scenes (BUSS) that comes with rigorously verified ground truth rotation.
\end{itemize}

\section{Related work}
\label{sec:rel_work}

Motion estimation methods can be classified into three groups: differential methods, discrete methods, and direct methods. Differential methods model the pixel displacements between two frames as instantaneous 3D velocities, while discrete methods model the pixel displacements as 3D translations and rotations.
 Direct methods typically avoid defining displacements explicitly, and are based on brightness-constancy constraints. Our method can be used in either the differential or discrete paradigms.

\paragraphCVPR{Differential methods}
These methods~\cite{Raudies2012ARA} (also known as instantaneous-time~\cite{Tian1996ComparisonOA} or continual methods) use visual motion field for estimating  camera motion, and are thereby well-suited for small motions. We start by reviewing methods based on the motion model proposed by Longuet-Higgins and Prazdny~\cite{LonguetHiggins1980TheIO}.

In \cite{Bruss1982PassiveN,MacLean1994RecoveryOE}, the authors formulate the problem to be independent of scene depth, and solve using nonlinear numerical optimization using a weighted bilinear constraint.
Kanatani~\cite{Kanatani1993RenormalizationFU} shows that these methods introduce bias. To remove the bias, Zhang and Tomasi~\cite{Zhang1999FastRA} propose an iterative method with an unweighted bilinear constraint that optimizes for translation using the Gauss-Newton method.
To avoid local minima of previous methods, Pauwels and Van Hulle~\cite{Pauwels2006OptimalIR} start with the weighted bilinear constraint
and gradually move to the unweighted bilinear constraint.
The authors in \cite{Rieger1985ProcessingDI,Prazdny2004EgomotionAR} leverage the fact that in areas of depth discontinuities, the difference of flows is due to translation.
Inspired by this work, Heeger proposes multiple subspace methods \cite{Heeger2004SubspaceMF, Jepson1994LinearSM, Heeger1990VisualPO} that also solve for translation first. The difference is that the solution is not an approximation and the sampled flow vectors do not need to be close to each other.

In~\cite{Perrone1992ModelFT}, Perrone 
tunes flow vector location-specific detectors to respond to different translation or rotation directions and speeds. He then adopts a voting scheme to determine the best-fitting rotation and translation.
A drawback is that this approach requires a huge number of templates to cover the 5 continuous dimensions.
In a subsequent work, Perrone \cite{Perrone1994AMO} proposes to first stabilize the gaze, reducing the number of dimensions to optimize. Lappe~\cite{Lappe1993ANN} proposes a biologically plausible implementation of~\cite{Heeger1990VisualPO}.

Methods in \cite{Maybank1985TheAV,Zhuang1988ASL,Kanatani20053DIO} use a differential version of the epipolar constraint.
\cite{Kanatani20053DIO,Zhuang1988ASL} propose algorithms to linearly solve for the continual fundamental matrix.

\paragraphCVPR{Discrete methods}
These methods do not make assumptions about frame-to-frame displacements, e.g., \cite{lin2010discrete,mainberger2008dense,kanatani2001comparing,triggs1999differential,baumela2000motion}.
The literature is vast. We list a few key works.

Most discrete methods use the epipolar constraint, \cite{LonguetHiggins1981ACA,Huang1989SomePO,Faugeras1993ThreedimensionalCV,ma2004invitation}, and can be split into two groups: calibrated and uncalibrated ones.
In both, most authors focus on deriving minimal solvers for RANSAC.
In the calibrated case, the essential matrix can be estimated using 5-point correspondences (see~\cite{Nistr2004AnES, Maybank1992TheoryOR:2004, Li2006FivePointME,Batra2007AnAF,Kukelova2008PolynomialES,Hartley2012AnEH}). Some authors proposed other minimal solvers for improving speed\footnote{The 5-point solver estimates up to 10 solutions for each sampling hypothesis, requiring up to 10 inlier counting per hypothesis. Non-minimal solvers get a single solution per hypothesis.}: \cite{Kahl199967point} derives 6- and 7-point solvers, and a DLT method (8-point algorithm) is presented in~\cite{ma2004invitation}.  \cite{Li2020RelativePE} uses $SE(3)$ invariances for constraining the minimal solvers. Others were proposed for constrained motions, such as \cite{Scaramuzza20111point,Bo20134points}. For the uncalibrated cases, again many authors proposed different solvers. In \cite{Hartley1997defence}, the author proposes an 8-point algorithm. \cite{Stewenius2008uncalibratedrelative,Kukelova2008GBgenerator} propose minimal 6-point algorithms for solving the relative pose with an unknown common focal length. \cite{Ding2022calibimu} explores the use of IMU for deriving a 4-point algorithm. In~\cite{Kuang2014relativedistortion}, the authors study problems with radial distortion.

Some authors focused on offering non-minimal solvers for fine estimates, such as~\cite{Briales2018ACG,Ding2021globalgrav,Zhao2022globalessential}. In \cite{Kneip2012FindingTE,Kneip2013DirectOO}, the authors propose a new epipolar constraint based on the coplanarity of epipolar plane normal vectors. \cite{Cimarelli2022RAUMVORA} avoids possible local minima by using an estimate of an unsupervised pose network. In~\cite{Garcia2021fast}, the authors use a robust loss function to detect and discard outliers.

\paragraphCVPR{Direct methods}
Instead of explicitly computing the optical flow, direct methods solve for camera motion using the brightness-consistency constraint equation (e.g., \cite{Silva1997RobustEE,Silva1996DirectEE,Fermller2005QualitativeE,Yuan2013ANM,Heel1990DirectEO}). Despite spatial-temporal gradient information, no closed-form solution exists, and strong assumptions must be made to simplify the problem. Horn and Weldon proposed several algorithms for cases of pure rotation, pure translation, or when depth is known \cite{Horn2004DirectMF}. Some works have shown that even when there exist rotation and translation, a solution can be found by considering the world as planer \cite{Horn2004DirectMF, Ai2021ADA, Ji2006A3S}, or by enforcing chirality constraint (the depth remains positive) \cite{Parameshwara2022DiffPoseNetDD, Barranco2021JointDE}. Direct methods suffer from changes of illumination, and they become extremely slow when run using a robust estimator framework to handle moving objects.

\paragraphCVPR{Robust motion estimation} To handle moving objects and noise in the (dense or sparse) correspondences, motion estimation methods are usually run within robust estimators.
Bideau et al.~\cite{Bideau2016ItsMA} develops a loss function to evaluate the quality of camera rotations with respect to optical flows, and perform gradient descent of this function in $SO(3)$. Unfortunately, local minima of the loss can be caused by similarity of rotation flows to translation and moving object flows, leading to poor solutions. To avoid this, one may do exhaustive search, such as in~\cite{Perrone1992ModelFT}, by discretizing rotation space into $C_{rot}$ 3D bins, and evaluate every rotation in time $\mathcal{O}(C_{rot})$. In general, large bins yield poor accuracy and smaller bins are too computationally expensive. Intuitively, this approach wastes large amounts of computation on very poor rotation candidates.
RANSAC~\cite{Fischler1981RANSAC} takes a different approach in which a random sample of flow vectors are evaluated for consistency with a particular motion model. This method works well when there are few outliers. However, with large numbers of outliers, such as in crowded street scenes, the required number of RANSAC iterations becomes too large, yielding very slow algorithms (see run time results for RANSAC algorithms). Thus, RANSAC is not viable when the percentage of outlier pixels is large. To tackle those challenges, we propose a generalization of the Hough Transform on $SO(3)$ that we describe in \cref{sec:voting}.

\section{Proposed approach}
\label{sec:method}
Our goal is to estimate the camera rotation between two frames, given $\{u_i, v_i, x_i, y_i\}$ where $(u_i, v_i)$ are optical flow vectors and $(x_i, y_i)$ are their respective coordinates in the image plane.
Consider an optical flow field $F$ caused purely by camera rotation, with no camera translation, moving objects, or noise. 
As we discuss in \cref{sec:compatible_rotations}, each flow vector in such a rotation field provides two constraints on the set of possible rotations, as shown in \cref{fig:intro}.  For a purely rotational optical flow field, the lines intersect in a single point, the rotation that causes the optical flow. 

However, in real-world video, optical flow is also affected by translation, moving objects and noise. In general, there exists no single rotation compatible with all the optical flow vectors. To estimate the rotation, we leverage the fact that the flows of distant points are mostly affected by rotation and thus behave approximately like 'rotation-only' flow vectors. The hypothesis is that these distant points will provide consistent evidence for a particular rotation, while other flow vectors, influenced by translation, scene geometry, moving objects, and noise, will not produce a consistent estimate of rotation. Thus, by accumulating evidence (or votes) for the rotation with strongest support, we can estimate the camera rotation. 

Of course, this highlights an important assumption of our method: we assume that camera translations between frames are small relative to distant points in the scene. This ensures that the flows of  distant scene points are well-modeled by rotations. Thus, our method is designed to work in outdoor scenes (or spacious indoor scenes, like arenas) where the translational camera motions are small relative to the most distant objects.

Our method can be considered a variation of the well-known Hough transform~\cite{osti_4746348}. The Hough transform attempts to find the hidden variable that could have generated as many observations as possible. Each observation is used to ``vote'' for the hidden variable values with which it is consistent. In our case, the observations are optical flow vectors (at each point in the image), and the hidden variable values are the possible rotations. This approach can be considered a ``robustification'' method, since it allows us to get good estimates in the presence of large numbers of ``outliers'', i.e., flows influenced by other factors (translation, moving objects, poor optical flow estimates). \\
The rest of this section is as follows: We review the perspective projection motion model and the Longuet-Higgins motion model in \cref{sec:perspective_projection} and \cref{sec:longuet_higgins}, and we derive the set of compatible rotations using both models; In \cref{sec:voting}, we introduce our Hough transform based voting scheme and compare the computational efficiency of our method to other robust methods.

\begin{figure}[t]
  \centering
   \includegraphics[width=0.4\linewidth]{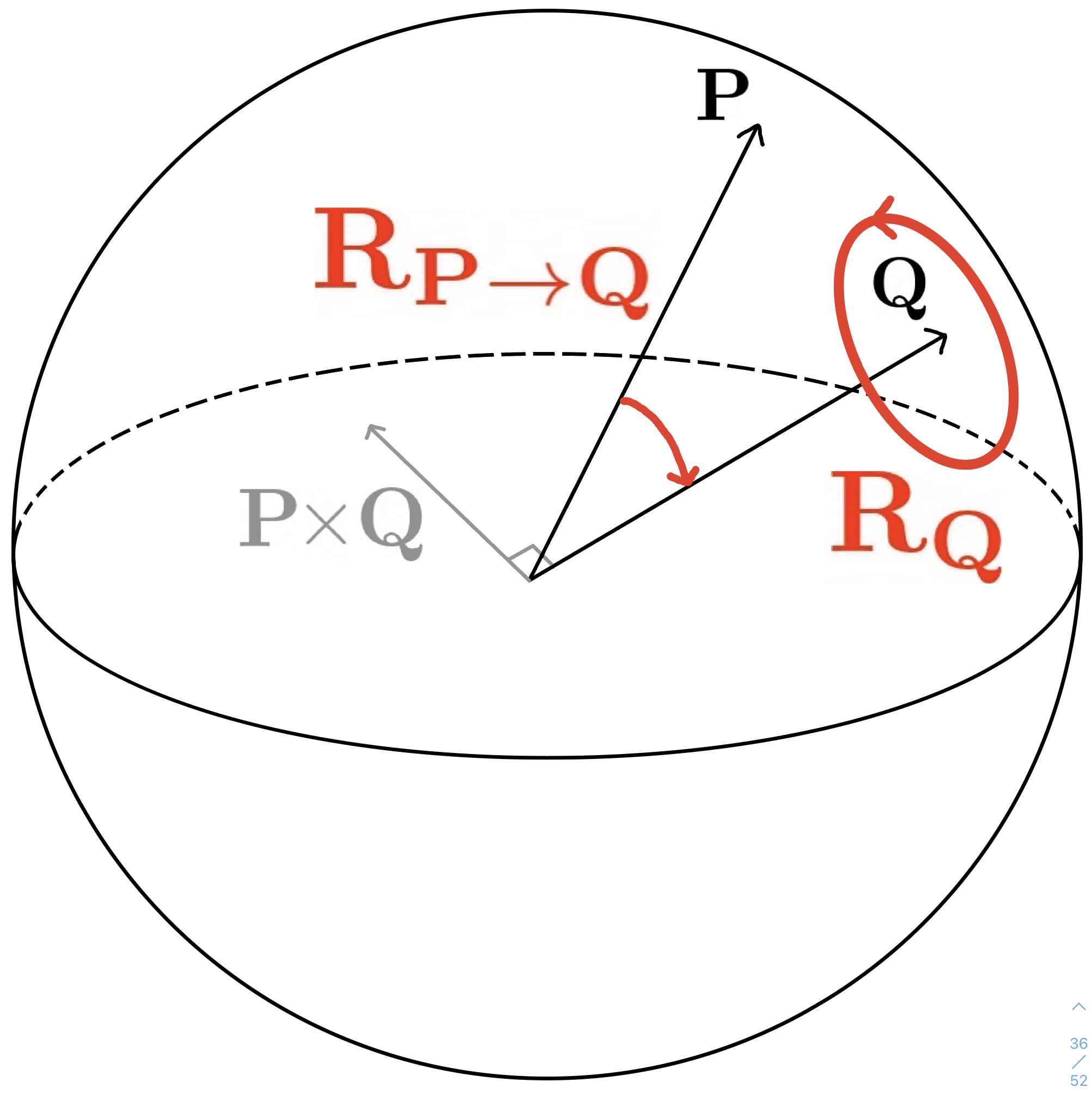}
    \caption{\textbf{Retrieving the set of rotations mapping $\mathbf{P}$ to $\mathbf{Q}$}. The set of all rotations that map $P$ to $Q$ (a one-dimensional submanifold of $SO(3)$) can be obtained by composing $\bold{R}_{P \rightarrow Q}$, a single rotation, with $\bold{R}_Q$, the set of all rotations about the vector $Q$.}
   \label{fig:sphere}
\end{figure}
\subsection{Compatible rotations}
\label{sec:compatible_rotations}
In this section, we discuss how to find the set of  rotations that can produce a specific optical flow vector that is only affected by camera rotation. Given that the space of 3D rotations $SO(3)$ is a 3D manifold  (rotations about the 3 axes) and that optical flow vectors have two degrees of freedom ($u$ and $v$), there is a one-dimensional set  of rotations with which any flow vector is compatible. We present two versions of our method, a discrete version using perspective projection, and a continual version using the Longuet-Higgins motion model.

\subsubsection{The perspective projection motion model}\label{sec:perspective_projection}
In this section we review classical materials on perspective projection, and we show how to compute the set of rotations that can produce a particular flow vector under perspective projection.
Consider a camera, aligned with world coordinates, that images a point with world coordinates $P$ and image coordinates $\mathbf{p}$. Now consider a rotation of the camera so that the world point in the new camera frame is given by camera coordinates $Q$ and image location $\mathbf{q}$. Because the magnitude of $P$ and $Q$ are the same (rotations do not change vector magnitudes), and the magnitudes of $P$ and $Q$ do not affect their projections onto the image, we can assume they both have unit magnitude. 

The set of all rotations that map $P$ to $Q$ (a one-dimensional submanifold of $SO(3)$) can be obtained by composing $\bold{R}_{P \rightarrow Q}$, a single rotation about $P\times Q$, with $\bold{R}^{\theta}_{Q}$, a rotation about the vector $Q$ by an angle $\theta$ that does not change the position of $Q$ in the camera's frame. This can be done for any angle $\theta$, generating a one-dimensional manifold of rotations (see \cref{fig:sphere}) that can be written as 
$$
\mathcal{R}(P,Q) = \{\bold{R}: \bold{R}=\bold{R}^{\theta}_{Q} \bold{R}_{P \rightarrow Q}, 0\leq \theta < 2\pi\}.
$$

\begin{figure}[t]
  \centering
   \includegraphics[width=0.667\linewidth]{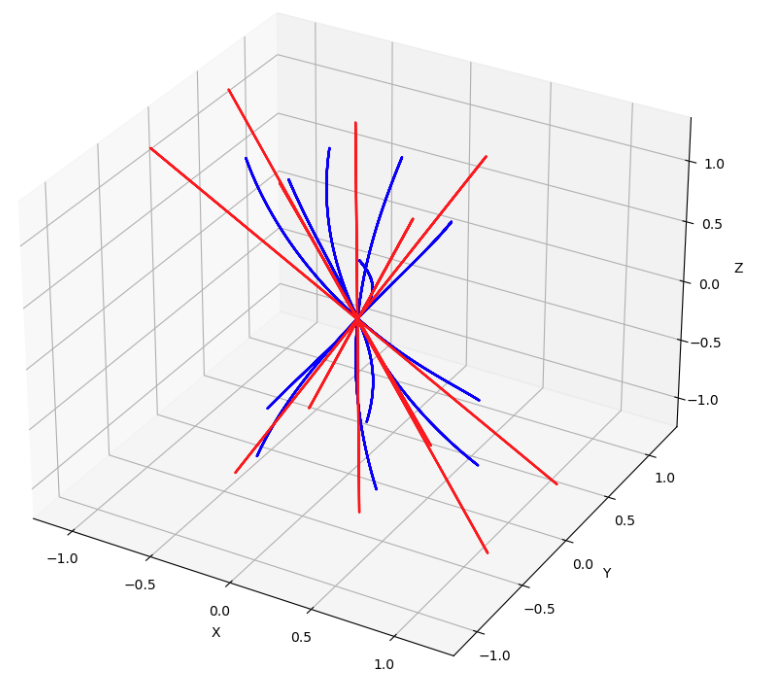}
    \caption{{\bf Longuet-Higgins vs.~perspective projection.} Each flow vector is compatible with a 1D manifold of rotations (axes in radians). Here, we show (partial) sets of compatible rotations using the Longuet-Higgins model (straight red lines) and (partial) sets of compatible rotations using perspective projection (blue curves.) Our algorithm can be used with either motion model.}
   \label{fig:perspective_vs_longuet_higgins}
\end{figure}

Given a set of rotations that can map $P$ to $Q$, we next discuss how to find rotations that map $\mathbf{p}$ to $\mathbf{q}$ in an image (i.e., the rotations that are compatible with optical flow vector $\mathbf{q}-\mathbf{p}$).
First, to get $P$ and $Q$, we take the inverse images of $\mathbf{p}$ and $\mathbf{q}$ (assuming that $P$ and $Q$ are unit vectors). The set of  rotations compatible with the flow vector $\mathbf{q}-\mathbf{p}$ is therefore $\mathcal{R}(P,Q)$.  To compute this from $P$ and $Q$ we define $\bold{R}_{P \rightarrow Q}$, which consists of an axis of rotation $P\times Q$, and the angle of rotation given by $\arccos(P\cdot Q)$. For $\bold{R}^{\theta}_{Q}$, the axis of rotation is simply $Q$, and the angle of rotation is any $\theta$ such that $0\leq \theta < 2\pi$. This one-dimensional family of rotations $\mathcal{R}(P,Q)$ is a curve in $SO(3)$ (blue curves in \cref{fig:perspective_vs_longuet_higgins}). Next, we show how the Longuet-Higgins model yields a slightly different set of compatible rotations.

\subsubsection{Using Longuet-Higgins motion model}\label{sec:longuet_higgins}

The Longuet-Higgins visual motion field model for static scenes~\cite{LonguetHiggins1980TheIO} defines an instantaneous motion field velocity (rate of change of position in the image) as 
\begin{equation}
\bold{v} %
 = 
 \underbrace{\begin{pmatrix}
  \frac{A}{f}xy  - Bf - \frac{B}{f}x^2 + Cy \\
  Af + \frac{A}{f}y^2 - \frac{B}{f}xy - Cx
 \end{pmatrix}}_{\bold{v_r}}
 +
 \underbrace{\begin{pmatrix}
  \frac{-fU+xW}{Z}\\
 \frac{-fV+yW}{Z}
 \end{pmatrix}}_{\bold{v_t}}.
 \label{eqn:flowdecomp}
\end{equation}
The motion field velocity $\bold{v}$ is represented as a sum of the 2D rotational velocity $\bold{v_r}$ and the 2D translational velocity $\bold{v_t}$. These in turn are defined as functions of the 3D translational velocities $U,V,W$, the 3D rotational velocities $A,B,C$,  the depth $Z$,  the image positions $x,y$ and the focal length $f$.
For motion caused only by rotation, we have, for a specific image location $(x,y)$, 
\begin{equation}
  \bold{v}(x,y) 
    =
  \begin{pmatrix}
  A\left(\frac{xy}{f}\right)  - B \left(\frac{f^2+x^2}{f}\right) + Cy \\
  A \left(\frac{f^2+y^2}{f}\right) - B\left(\frac{xy}{f}\right) - Cx
 \end{pmatrix}.
 \label{eqn:eqsystem}
\end{equation}%
These equations lead to a 1D manifold of solutions, a line $l$ at the intersection of two planes defined by the two equations in \cref{eqn:eqsystem}. {\bf The simple form of this 1D manifold (a straight line) allows a very fast implementation of the Hough transform, as described in \cref{sec:voting}.}

Let $\bold{n}_u$ and $\bold{n}_v$  be  normal vectors to these planes:
\begin{equation}
 \bold{n_u} = \left[\frac{xy}{f}, -\frac{f^2+x^2}{f}, y\right], \
\bold{n_v} = \left[\frac{f^2+y^2}{f}, -\frac{xy}{f}, -x\right].
\label{eqn:line_direction}
\end{equation}
The line $l$ defined by the intersection of the two planes has direction $\mathbf{d} = \bold{n_u} \times \bold{n_v}$. 
By simple algebra, it can be shown that the $z$ component of $\bold{d}$ can't be $0$, which implies that the line $l$ can't be co-planar to the plane $C=0$. Therefore, we can complete the definition of $l$ by  finding its intersection with the plane $C=0$, by setting $C=0$ in \cref{eqn:eqsystem}.
Notice that the direction of $l$ (given by the vector $\bold{d}$) is independent of the optical flow vector. Only the intercept depends on the flow vector. Therefore, one can precompute line directions for each image location, and only find the intercept at run time, resulting in major efficiency gains. \Cref{fig:perspective_vs_longuet_higgins} shows  the 1D manifolds of compatible rotations produced by perspective projection and the Longuet-Higgins motion model. A comparison of the accuracy and the run time of the two approaches can be found in \cref{sec:ablation_bin}.  In the subsequent sections of the paper, we will report results of our method using the Longuet-Higgins motion model.
\subsection{Voting Scheme}
\label{sec:voting}

We discretize the 1D manifold of solutions we get from \cref{sec:compatible_rotations} into rotation votes.
Unlike the original Hough transform, we do not create an accumulator, but make a list of compatible rotation votes, and find the mode of the list, alleviating the need for a 3-dimensional accumulator in memory. In summary, our approach allows dense sampling of $SO(3)$ while maintaining rapid execution.
Our method's speed depends on the number of flow vectors $C_{OF}$ used in voting and the number of points sampled per 1D manifold of compatible rotations. We sample about $\sqrt[3]{C_{rot}}$ points per 1D manifold, the approximate number of bins intersected by each line. Thus, our total number of `votes' and hence our complexity is $\mathcal{O}(C_{OF}\sqrt[3]{C_{rot}})$.
\section{Dataset}
\label{sec:dataset}
\begin{figure}[t]
  \centering
   \includegraphics[width=1\linewidth]{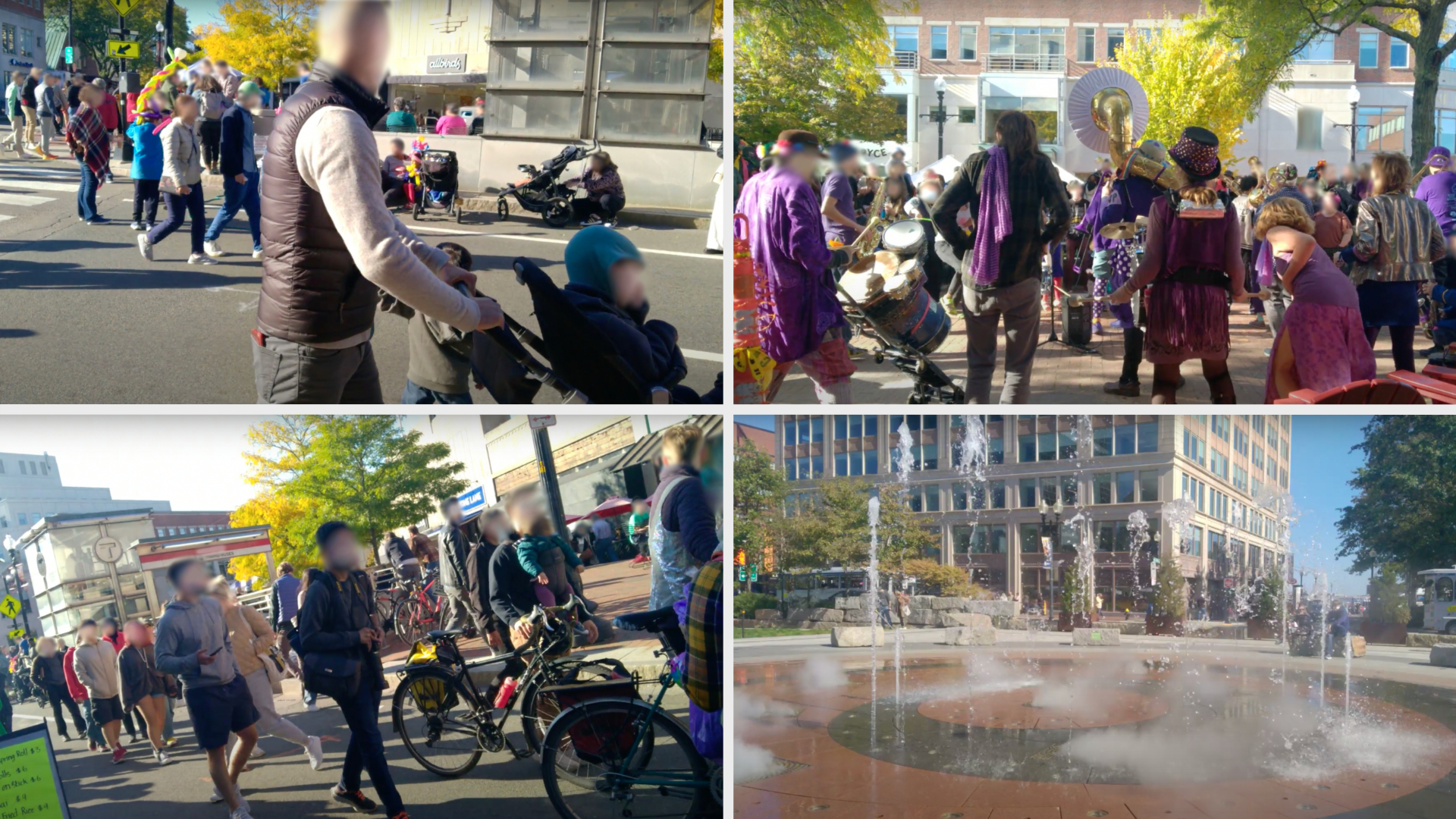}
   \caption{\textbf{BUSS dataset}. Example frames from our BUSS dataset. The sequences are recorded in different scenes and have a diverse set of camera motion.}
   \label{fig:buss-frames}
\end{figure}
We introduce \textit{BUsy Street Scenes} (BUSS), a challenging dataset of video sequences taken from a handheld mobile phone (an OPPO A5 2020 smartphone, rear camera) in crowded city streets with synchronized inertial measurement unit (IMU) data. The goal of the dataset is to evaluate the robustness of camera rotation estimation algorithms in dense and dynamic scenes with many moving objects and complex camera motion. The dataset composes 17 video sequences of about 10 seconds each at 30fps in full HD resolution (1920x1080) RGB. We used the \emph{Android Open-Camera Sensor} app to synchronously record video and angular rate from the phone's MEMS gyroscopes (at 400Hz) and then generated the rotation ground truth using the method we discuss in \cref{sec:ground_truth_calculation}. To meet strict privacy standards, videos are only captured in public places, and faces and other personally identifiable information (PII) is blurred. Along with the anonymized video frames, we also provide optical flow for all sequences computed with RAFT~\cite{Teed2020RAFTRA}. All sequences show highly dynamic scenes (see \cref{fig:buss-frames}).

\begin{table}[t]
  \caption{
\textbf{Dataset comparison}. Comparison of our proposed BUSS dataset to other relevant datasets.
}
  \centering
\resizebox{1\linewidth}{!}{\setlength{\tabcolsep}{5.0pt}\begin{NiceTabular}{@{}lcccc@{}}[colortbl-like]
  \toprule
 & \thead{BUSS \\ (ours)} & \thead{IRSTV \cite{Servieres2021VisualAV}} & \thead{Cambridge\\Landmarks \cite{Kendall_2015_ICCV}} & \thead{KITTI \cite{Geiger2013IJRR}} \\
   \midrule
    \rowcolor{Gray!20}\thead{Year} &
    \makecell{2022} &
    \makecell{2021} &
    \makecell{2016} &
    \makecell{2012} \\ 
    \rowcolor{white!20}\thead{Platform} &
    \makecell{Hand held} &
    \makecell{Hand held} &
    \makecell{Hand held} &
    \makecell{Car} \\ 
    \rowcolor{Gray!20}\thead{Scene} &
    \makecell{Outdoors} &
    \makecell{In/outdoors} &
    \makecell{Outdoors} &
    \makecell{Outdoors} \\ 
    \rowcolor{white!20}\thead{\% of moving objects} &
    \makecell{Very high} &
    \makecell{Very low} &
    \makecell{Low} &
    \makecell{Very low} \\
    \rowcolor{gray!20}\thead{Anno. freq.} &
    \makecell{30Hz} &
    \makecell{20Hz} &
    \makecell{2Hz} &
    \makecell{10Hz} \\
    \rowcolor{white!20}\thead{Rot. GT} &
    \makecell{IMU} &
    \makecell{IMU} &
    \makecell{SfM} &
    \makecell{IMU}\\
    \rowcolor{gray!20}\thead{Baseline} &
    \makecell{Small} &
    \makecell{Small} &
    \makecell{Large} &
    \makecell{Large} \\
    \rowcolor{white!20}\thead{Num. frames} &
    \makecell{5,504} &
    \makecell{7,800} &
    \makecell{10,929} &
    \makecell{43,552}\\
    \bottomrule
  \end{NiceTabular}}
  \label{tab:datasets}
\end{table}
\subsection{Ground truth calculation}
\label{sec:ground_truth_calculation}
The BUSS ground truth was estimated using the angular rate measurements recorded simultaneously with the video. The ground truth rotation at frame $f_t$ represents the forward rotation from the video frame $f_t$ to the immediate next frame $f_{t+1}$. To get the rotation between two frames, we numerically integrate angular rate measurements~\cite{5975346}.

To assess the reliability of our dataset's rotational ground truth, we compared the measurements of the OPPO with the measurements of a different phone (iPhone 12 mini) with the phones bound to the same rigid surface. Comparing two gyroscope sensor models gives us strong confidence in data correctness since it is highly unlikely that the two phones (with different hardware) agree on erroneous measurements. After recording gyroscope data simultaneously from both phones, we corrected for temporal and spatial misalignment. We synchronized the internal clocks of the two phones by finding the time offset that minimized disagreement error. For spatial misalignment, we corrected for the relative orientation $\bold{R}$ between the two gyroscopes using the Kabsch algorithm on the rotation velocity vectors~\cite{Kabsch1976ASF}. The 
average frame-to-frame (at 30 fps) rotation error between the two phones is $0.014^{\circ}$. This is an order of magnitude smaller than the errors from state-of-the-art methods in a similar setting. This validates the choice of using a gyroscope to generate the ground truth of the dataset. See additional details in supplementary material.

\subsection{Comparison to existing datasets}
\label{sec:dataset_comparison}

Our proposed BUSS dataset has three key properties that are not found simultaneously in any publicly available dataset:
(a) it is recorded with a handheld camera, introducing highly variable camera motions (b) it contains highly dynamic scenes, and (c) it has high frequency, accurate and synchronized rotation ground truth. %
The IRSTV dataset~\cite{Servieres2021VisualAV} does not have property (b) because the number of moving objects is sparse. The Cambridge Landmarks dataset~\cite{Kendall_2015_ICCV} contains some sequences with dynamic scenes, but the ground truth rotations are only given at 2 FPS. The popular KITTI dataset~\cite{Geiger2013IJRR} has few moving vehicles and pedestrians per frame and the camera is mounted on a vehicle, so the dataset is lacking all three properties. 
The comparison of the main characteristics of our dataset with three other publicly available datasets can be found in \cref{tab:datasets}.

\section{Experiment}
\label{sec:experiments}

We evaluate our method for frame-to-frame rotation estimation on our proposed BUSS dataset and on IRSTV \cite{Servieres2021VisualAV}.  The properties of both datasets are described in \cref{sec:dataset_comparison}.

\subsection{Evaluation Metrics}

To evaluate the rotation estimation accuracy, we use Average Angular Error (AAE) which computes the angle of rotation between the ground-truth rotation and estimated rotation. Let $f_{i, j}$ be the $j^{th}$ frame in the $i^{th}$ sequence. Let $\bold{R}_{i, j}$ be the ground-truth rotation between frame $f_{i, j}$ and frame $f_{i, j+1}$, and $\hat{\bold{R}}_{i, j}$ be the estimated rotation between the same two frames. Then, the average angular error is given by
\begin{align}
    AAE = \frac{1}{\sum_{i=1}^{N}M_i} \sum_{i=1}^{N}\sum_{j=1}^{M_i}
               \theta(\hat{\bold{R}}_{i, j}\bold{R}_{i, j}^{-1}) ,  
\end{align}
where $N$ is the number of sequences, $M_i$ is the number of optical flows for the sequence $i$ and $\theta(\cdot)$  is the magnitude of rotation resulting from $\hat{\bold{R}}_{i, j}\bold{R}_{i, j}^{-1}$.
\label{eq:aae}

\subsection{Implementation Details}
All methods, including ours, are run on an Intel Xeon CPU. For the continual baselines, we used the MATLAB implementations and the RANSAC parameters provided in~\cite{Raudies2012ARA}. For the discrete baselines, we used the implementations and the RANSAC parameters from OpenGV~\cite{Kneip2014OpenGVAU}. To offer a fair run time comparison against other continual baselines, our method is also implemented in MATLAB. We use a bin size of 0.057 degrees (see the ablation in \cref{sec:ablation_bin}), and we search over rotations between -4 and 4 degrees.

For the methods that require the optical flow, including ours, we first resize the video frames to a size of 480x270. We then compute the optical flow  using RAFT~\cite{Teed2020RAFTRA} on a GeForce GTX 1080Ti GPU. RAFT offers a good trade-off between performance and speed. We then resized the optical flow to 32x18 by sampling the optical flow vectors on a regular grid with stride 15. An ablation on the spatial sampling rate can be found in \cref{sec:ablation_spatial}.

For the methods based on feature correspondences, we extract 3000 SIFT descriptors from the full-resolution 1920x1080 video frames. We match them using a standard brute-force matcher.

\subsection{Results}
\label{sec:results}
We compare the frame-to-frame rotation estimation from our Longuet-Higgins method against several continual baselines: Bruss\&Horn (B\&H)~\cite{Bruss1982PassiveN}, Heeger\&Jepson (H\&J)~\cite{Heeger2004SubspaceMF}, Kanatani (Kan)~\cite{Kanatani20053DIO}, Lappe\&Rauschecker (L\&R)~\cite{Lappe1993ANN}, Pauwels\&Van Hulle (P\&V)~\cite{Pauwels2006OptimalIR}, Zhang\&Tomasi (Z\&T)~\cite{Zhang1999FastRA}, and discrete baselines: Kneip (Kne)~\cite{Kneip2013DirectOO} and Nist\'er also known as the 5-points algorithm (Nis)~\cite{Nistr2004AnES}. 
In addition of running the continual methods using all the optical flow vectors, we also run all continual methods, except Lappe\&Rauschecker~\cite{Lappe1993ANN} and ours which are robust to moving objects by design, in RANSAC for 1, 25, 100 and 500 RANSAC iterations. For the discrete baselines, we ran each method for 500, 5000 and 50000 iterations. To quantify the amount of rotation in both datasets, we also include the zero baseline as reference.

\begin{figure}[h!]
  \centering
   \includegraphics[width=1\linewidth]{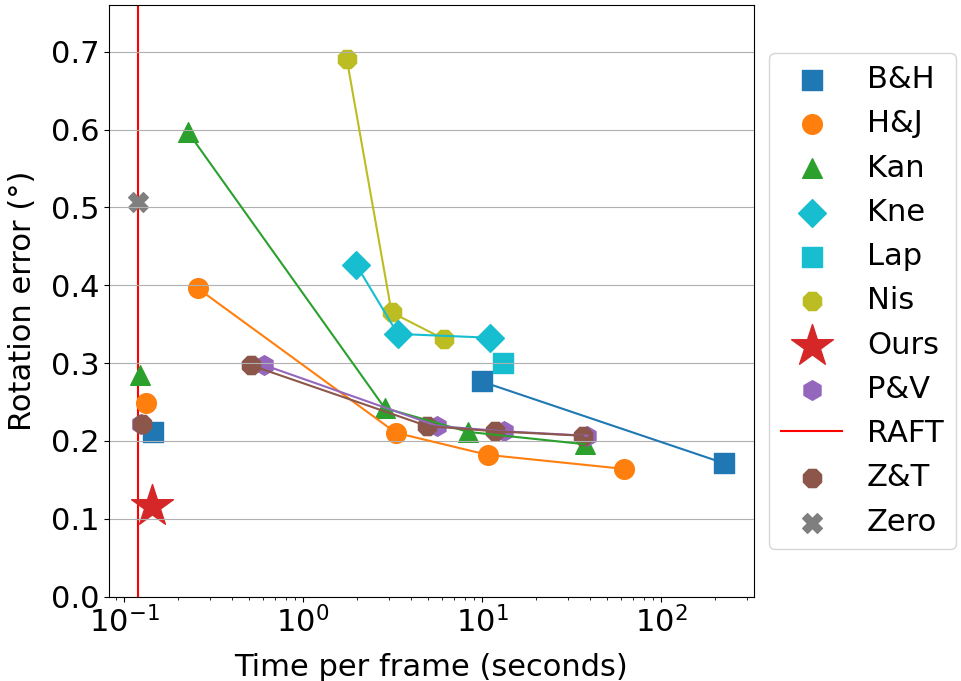}
    \\

\vspace{.3em}
    \caption{\textbf{Rotation error vs.~run time on BUSS.} Methods run with RANSAC appear on a line, with different numbers of RANSAC iterations at each point. Standalone points do not use RANSAC. The run time of the continual methods includes the run time of the optical flow computation.}
   \label{fig:buss_plot_res}
\end{figure}

\paragraphCVPR{Results on BUSS} The results on the BUSS dataset clearly illustrate the strength of our approach. \Cref{tab:buss_quant_results} reports the numerical results, and \cref{fig:buss_plot_res} shows the rotation error vs. run time. Our method is almost 50\% more accurate than comparably fast methods. Due to the highly dynamic nature of the BUSS dataset, RANSAC significantly improves the accuracy of the other methods (ranging from 66\% to 30\% improvement). Yet, even with the improvement gained from RANSAC, our method outperforms the second-best method by 25\% while being more than 400 times faster. The standard error of the mean is smaller than 1.3\% for our method, and smaller than 7\% for the other methods.

\begin{table}[h!]

  \centering
    
\resizebox{1\linewidth}{!}{\setlength{\tabcolsep}{2pt}\begin{NiceTabular}{@{}lccccc|ccccc@{}}[colortbl-like]
\toprule
& \multicolumn{5}{c}{\thead{Rotation err. (°)}} & \multicolumn{5}{c}{\thead{Time per frame (seconds)}}  \\ \midrule
\rowcolor{Gray!20} & \multicolumn{10}{c}{\it Continual methods}\\
\rowcolor{Gray!20} \thead{\# iters.} & N/A  & 1 & 25 & 100 & 500 & N/A & 1 & 25 & 100 & 500 \\ 
\rowcolor{Gray!20}\multirow{1}{*}{\thead{B\&H~\cite{Bruss1982PassiveN}}} & 0.21 & 0.28 & 0.17 & ---$^{*}$ & ---$^{*}$ & 0.14 & 9.92 & 226.01 & ---$^{*}$ & ---$^{*}$  \\
\rowcolor{Gray!20}\multirow{1}{*}{\thead{H\&J~\cite{Heeger2004SubspaceMF}}} & 0.25 & 0.40 & 0.21 & 0.18 & 0.16 & 0.13 & 0.26 & 3.30 & 10.72 & 62.09  \\
\rowcolor{Gray!20}\multirow{1}{*}{\thead{Kan~\cite{Kanatani20053DIO}}} & 0.28 & 0.60 & 0.24 & 0.21 & 0.20 & 0.12 & 0.23 & 2.86 & 8.37 & 37.54  \\
\rowcolor{Gray!20}\multirow{1}{*}{\thead{L\&R~\cite{Lappe1993ANN}}} & 0.30 & --- & --- & --- & --- & 13.07 & --- & --- & --- & ---  \\
\rowcolor{Gray!20}\multirow{1}{*}{\thead{P\&V~\cite{Pauwels2006OptimalIR}}} & 0.22 & 0.30 & 0.22 & 0.21 & 0.21 & 0.12 & 0.61 & 5.62 & 13.27 & 38.76  \\
\rowcolor{Gray!20}\multirow{1}{*}{\thead{Z\&T~\cite{Zhang1999FastRA}}} & 0.22 & 0.30 & 0.22 & 0.21 & 0.21 & 0.13 & 0.51 & 4.93 & 11.82 & 36.50 \\
\rowcolor{Gray!20}\multirow{1}{*}{\thead{Ours}} & 0.12 & --- & --- & --- & --- & 0.14 & --- & --- & --- & ---  \\ 
\midrule
& \multicolumn{10}{c}{\it Discrete methods}\\
\thead{\# iters.} &
 & 500 & 5K & 50K & & & 500 & 5K & 50K &  \\ 
\multirow{1}{*}{\thead{Kneip~\cite{Kneip2013DirectOO}}} & & 0.69 & 0.36 & 0.33 & & & 1.63 & 3.00 & 6.04 &  \\ 
\multirow{1}{*}{\thead{Nist\'er~\cite{Nistr2004AnES}}} & & 0.43 & 0.34 & 0.33 & & & 1.86 & 3.26 & 10.91 &  \\ 
\bottomrule
\end{NiceTabular}}
\caption{ \textbf{Quantitative results on the BUSS dataset}. We compare the frame-to-frame rotation error and the run time of our method with multiple other baselines for multiple number of RANSAC iterations. Experiments where the number of iterations is "N/A" means that the experiments have been run without RANSAC.\\$^{*}$\textit{The run time is too long to run the experiment.}}
\label{tab:buss_quant_results}
\end{table}

\begin{figure}[h]
  \centering
   \includegraphics[width=1\linewidth]{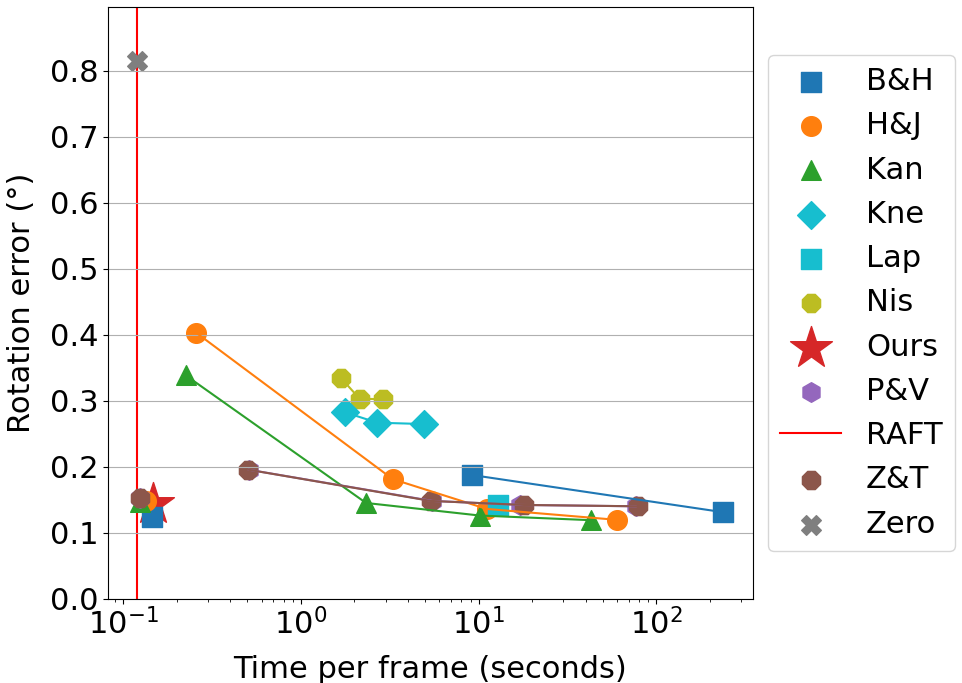}
\\
\caption{\textbf{Rotation error vs. run time on IRSTV.} Methods connect by lines use RANSAC. Standalone points do not.
}
\label{fig:irstv_plot_res}
\end{figure}

\paragraphCVPR{Results on IRSTV} The results for the IRSTV dataset are reported in \cref{tab:irstv_quant_results}. We show the plot of the rotation error vs. run time in \cref{fig:irstv_plot_res}. Our method is on par with the other methods with respect to accuracy and speed. Our method has a rotation error of 0.14° while operating at 0.15 seconds per frame. %
 Due to the fact that IRSTV is mostly composed of static scenes, running continual methods with RANSAC only marginally improve the results while increasing the run time significantly.

\begin{table}[h!]
  \centering
\resizebox{1\linewidth}{!}{\setlength{\tabcolsep}{2.0pt}\begin{NiceTabular}{@{}lccccc|ccccc@{}}[colortbl-like]
\toprule
& \multicolumn{5}{c}{\thead{Rotation err. (°)}} & \multicolumn{5}{c}{\thead{Time per frame (seconds)}}  \\ \midrule
\rowcolor{Gray!20} & \multicolumn{10}{c}{\it Continual methods}\\
\rowcolor{Gray!20} \thead{\# iters.} & N/A  & 1 & 25 & 100 & 500 & N/A & 1 & 25 & 100 & 500 \\ 
\rowcolor{Gray!20}\multirow{1}{*}{\thead{B\&H~\cite{Bruss1982PassiveN}}} & 0.12 & 0.19 & 0.13 & ---$^{*}$ & ---$^{*}$ & 0.15 & 9.15 & 238.45 & ---$^{*}$ & ---$^{*}$  \\
\rowcolor{Gray!20}\multirow{1}{*}{\thead{H\&J~\cite{Heeger2004SubspaceMF}}} & 0.15 & 0.40 & 0.18 & 0.14 & 0.12 & 0.13 & 0.26 & 3.28 & 11.09 & 59.64  \\
\rowcolor{Gray!20}\multirow{1}{*}{\thead{Kan~\cite{Kanatani20053DIO}}} & 0.15 & 0.34 & 0.14 & 0.13 & 0.12 & 0.12 & 0.22 & 2.32 & 10.11 & 43.04  \\
\rowcolor{Gray!20}\multirow{1}{*}{\thead{L\&R~\cite{Lappe1993ANN}}} & 0.14 & --- & --- & --- & --- & 12.78 & --- & --- & --- & ---  \\
\rowcolor{Gray!20}\multirow{1}{*}{\thead{P\&V~\cite{Pauwels2006OptimalIR}}} & 0.15 & 0.20 & 0.15 & 0.14 & 0.14 & 0.12 & 0.51 & 5.48 & 17.09 & 76.47  \\
\rowcolor{Gray!20}\multirow{1}{*}{\thead{Z\&T~\cite{Zhang1999FastRA}}} & 0.15 & 0.20 & 0.15 & 0.14 & 0.14 & 0.12 & 0.51 & 5.36 & 17.95 & 78.69 \\
\rowcolor{Gray!20}\multirow{1}{*}{\thead{Ours}} & 0.14 & --- & --- & --- & --- & 0.15 & --- & --- & --- & ---  \\ 
\midrule
& \multicolumn{10}{c}{\it Discrete methods}\\
\thead{\# iters.} &
 & 500 & 5K & 50K & & & 500 & 5K & 50K &  \\ 
\multirow{1}{*}{\thead{Kne~\cite{Kneip2013DirectOO}}} & & 0.28 & 0.27 & 0.26 & & & 1.64 & 2.57 & 4.82 &  \\ 
\multirow{1}{*}{\thead{Nis~\cite{Nistr2004AnES}}} & & 0.33 & 0.30 & 0.30 & & & 1.56 & 2.02 & 2.77 &  \\ 
\bottomrule
\end{NiceTabular}}
\caption{\textbf{Quantitative results on IRSTV dataset}. We compare the frame-to-frame rotation error and the run time of our method with multiple other baselines for multiple number of RANSAC iterations. Experiments where the number of iterations is "N/A" means that the experiments have been run without RANSAC.\\
$^{*}$\textit{The run time is too long to run the experiment.}}
 \label{tab:irstv_quant_results}
\end{table}

The results on IRSTV and BUSS show the robustness of our method to moving objects. While the rotation error of our proposed algorithm stays comparable across the two datasets, the rotation errors of the baselines increase on the BUSS dataset. It's worth noting that continual methods perform significantly better than the discrete methods on both datasets, which suggests that discrete methods are more susceptible to noise. Additionally, the Zero baseline error is more significant ($\approx 0.8$°) on the IRSTV dataset than on the BUSS dataset ($\approx 0.5$°) due to IRSTV having a lower frame rate. This also explains why our method performs slightly worse on IRSTV than on BUSS.

\subsection{Robustness to moving and close objects}

\begin{figure}[h!]
    \centering
    \includegraphics[width=.96\linewidth]{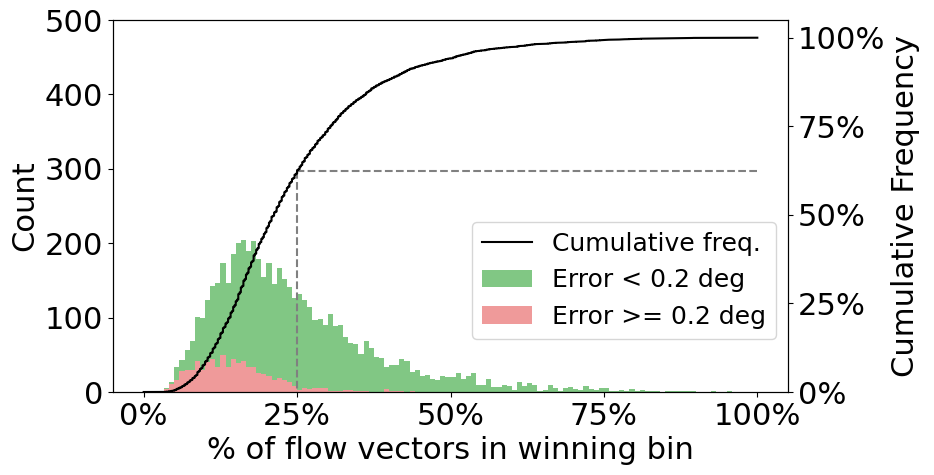} %
    \caption{Percentage of flow vectors that has voted for the winning rotation bin on the BUSS dataset. E.g., the dotted line shows that for 62\% of the optical flows, less than 25\% of the flow vectors are in the winning bin.}
    \label{fig:translation_percent_flow}
\end{figure}

In this section, we investigate the proportion of pixels in the frames needed to be far away. In the case of pure rotation, the winning rotation bin receives votes from all flow vectors. \cref{fig:translation_percent_flow} shows the percentage of flow vectors in the winning bin for the BUSS dataset. 62\% of the optical flows have less than 25\% of the flow vectors in the winning bin, with the majority resulting in small errors ($<$0.2deg). This shows that our algorithm is highly effective even when most of the flow vectors are affected by a translation or by moving objects.

\section{Ablation Studies}
\label{sec:ablation}

We compare motion models (perspective vs.~Longuet-Higgins), different quantizations of rotation space, and the spatial sampling rates of optical flow.

\begin{figure}[t]
  \centering
   \includegraphics[width=0.96\linewidth]{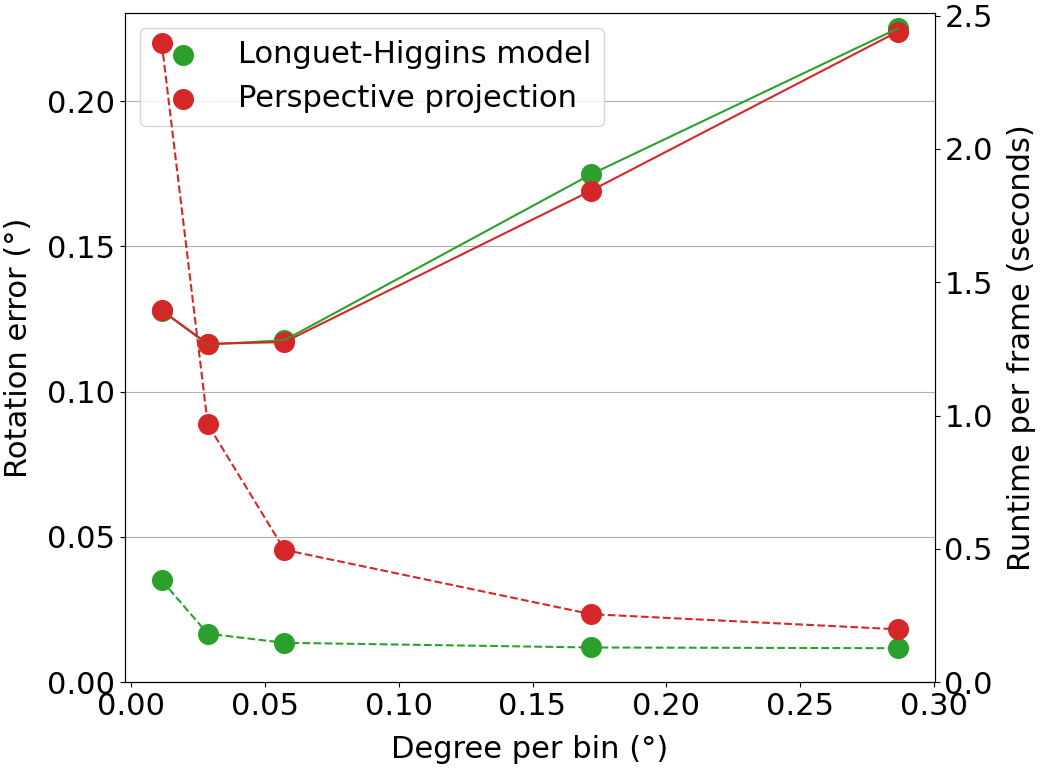}
   \caption{\textbf{Performance on BUSS as a function of bin size.}  Our method's accuracy (continuous line) and run time (dashed line) with perspective projection and Longuet-Higgins. The methods have similar accuracy but Longuet-Higgins is much faster.}
   \label{fig:bin_size}
\end{figure}

\paragraphCVPR{Varying bin sizes on $SO(3)$}
\label{sec:ablation_bin}
\cref{fig:bin_size} compares  rotation error and  run time for our two approaches on BUSS. Both methods demonstrate similar rotation accuracy regardless of  bin size. However Longuet-Higgins is much faster for small bins.
There is a sweet spot for bin size. If bins are too small, noise in optical flow  prevents the 1d manifold of compatible rotations from intersecting the correct bin.  But making bins bigger increases error when picking the correct bin. Since the rotation estimate is the bin's center, the maximum error when choosing the correct bin is $\left(\sqrt{3}/2\right)s$, for bin size $s$.

\begin{figure}[t]
  \centering
   \includegraphics[width=0.91\linewidth]{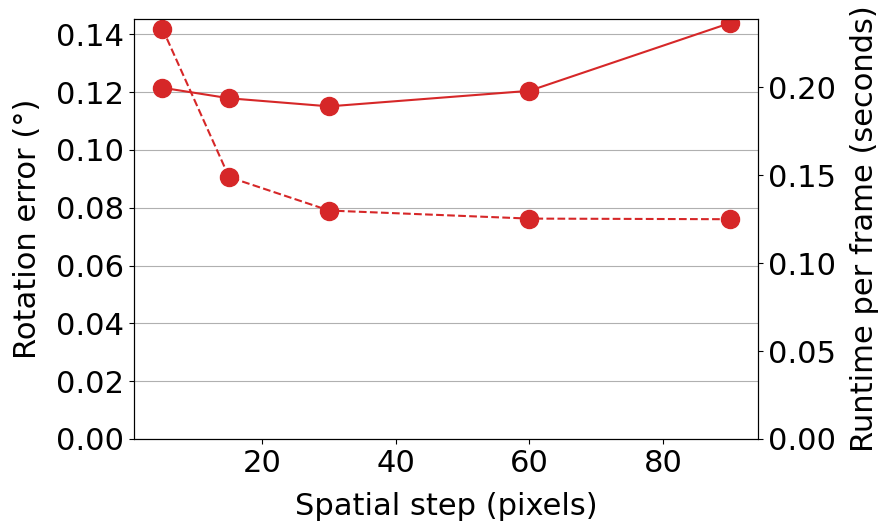}
   \caption{\textbf{Performance on BUSS as a function of spatial step size}. Our error (continuous line) and run time (dashed line) for different spatial step values. A spatial step value of $n$ means that we sample  flow vectors every $n$ pixels.}
   \label{fig:spatial_sampling}
\end{figure}

\paragraphCVPR{Robustness to spatial sampling}
\label{sec:ablation_spatial}
We sample  optical flow vectors along a regular grid. \cref{fig:spatial_sampling} shows our model's robustness to spatial sampling step sizes on BUSS. The rotation error remains between 0.11 degrees and 0.13 degrees for step sizes ranging from 1 to 80. This allows subsampling optical flow and reducing run time. When flow sampling becomes too sparse, rotation error increases due to overexposure to potentially noisy flow vectors. Surprisingly, rotation error also slightly increases when flow sampling becomes too dense. We hypothesize that this is because far away points are spatially distributed on the frames, while objects, that could potentially exhibit motions that are coherent to a rotation, are generally well-bounded in space. Therefore, after diminishing the sampling rate, we still benefit from far away points across the frame, while reducing flows sampled from the same object. 

\section{Conclusion}
\label{sec:conclusion}
We introduce a novel generalization of the Hough transform on $SO(3)$ to find the camera rotation most compatible with optical flow in highly dynamic scenes. Our method is inherently robust, and doesn't need RANSAC, which significantly improves the speed over existing methods. In presence of moving objects, our method reduces the error by almost 50\% over the next best method for the same run time, while performing similarly in static scenes. Additionally, we propose a challenging new dataset BUSS that consists of 17 video sequences in crowded, real-world scenes.

{\small
\bibliographystyle{ieee_fullname}
\bibliography{egbib}
}

\end{document}